# Go Beyond Multiple Instance Neural Networks: Deep-learning Models based on Local Pattern Aggregation

Linpeng Jin

Hangzhou Normal University, Hangzhou, 311121, China

**Abstract:** Deep convolutional neural networks (CNNs) have brought breakthroughs in processing clinical electrocardiograms (ECGs), speaker-independent speech and complex images. However, typical CNNs require a fixed input size while it is common to process variable-size data in practical use. Recurrent networks such as long short-term memory (LSTM) are capable of eliminating the restriction, but suffer from high computational complexity. In this paper, we propose local pattern aggregation-based deep-learning models to effectively deal with both problems. The novel network structure, called LPANet, has cropping and aggregation operations embedded into it. With these new features, LPANet can reduce the difficulty of tuning model parameters and thus tend to improve generalization performance. To demonstrate the effectiveness, we applied it to the problem of premature ventricular contraction detection and the experimental results shows that our proposed method has certain advantages compared to classical network models, such as CNN and LSTM.

**Key Words:** Deep Learning, Convolutional Neural Networks, Multiple Instance Neural Network

## 1. Introduction

In recent years, deep learning has made major breakthroughs in hard Artificial Intelligence tasks, such as large-scale image classification [1], speaker-independent speech recognition [2] and clinical electrocardiogram (ECG) analysis [3]. Among a variety of deep-learning models, Convolutional Neural Networks (CNNs) [4] are largely responsible for the success and play an irreplaceable part in many pattern recognition systems today.

A typical CNN is composed of alternating layers of convolution and pooling (namely convolutional unit), and fully-connected layers that follow. The convolutional units accept arbitrary input sizes, but they produce feature maps of variable sizes. Due to the fixed-length input defined by fully-connected layers, the CNN can only process fixed-size input data. However, it is impossible to have the same size of data in any given tasks, thus intuitive approaches fitting the input data to the fixed size are commonly used, such as cropping [1], padding [5] and warping [6]. Unfortunately, they create their own problems: (1) cropping can remove the contaminated data but also feature information for classification purposes; (2) padding can result in excessive use of computational resources and decrease the importance of distinguishing features, especially when the difference in input sizes is remarkable; (3) warping may give rise to unwanted geometric distortion, resulting in compromised recognition accuracy. In view of the above-mentioned facts, scholars equipped traditional CNNs with adaptive pooling strategies, enabling them to process input data of arbitrary sizes. Global Max Pooling (GMP) [7] and Global Average Pooling (GAP) [8] generate one value for each feature map (with arbitrary size) using the corresponding operations respectively, while Spatial Pyramid Pooling (SPP) [9] extends GMP/GAP and maps each feature map into a fixed-length vector using multi-level spatial bins. Nevertheless even with adaptive pooling strategies, CNNs will not perform well if input sizes vary greatly. The good news is that with the help of attentional mechanism (AM) [10], convolutional units together with recurrent structures represented by Long Short-Term Memory (LSTM) [11] and Gated Recurrent Unit (GRU) [12] can effectively deal with this problem. However, these sophisticated deep-learning models suffer from high computational complexity, and extra optimization tricks such as Dropout [13] and Batch Normalization [14] need to be used in the training/testing stage, which in turn further increase computational burden.

In this paper, we introduce a generic local pattern aggregation-based deep-learning model (LPANet), which enables CNNs and their variants to accept variable input sizes and overcomes the problems discussed above to some extent, such as decreased performance due to remarkable difference in input sizes, high computation complexity and difficulty of tuning model parameters.

## 2. Model Description

### 2.1 Overall Framework

Fig.1 shows the overall framework of our proposed LPANet and the technical details are described as follows:

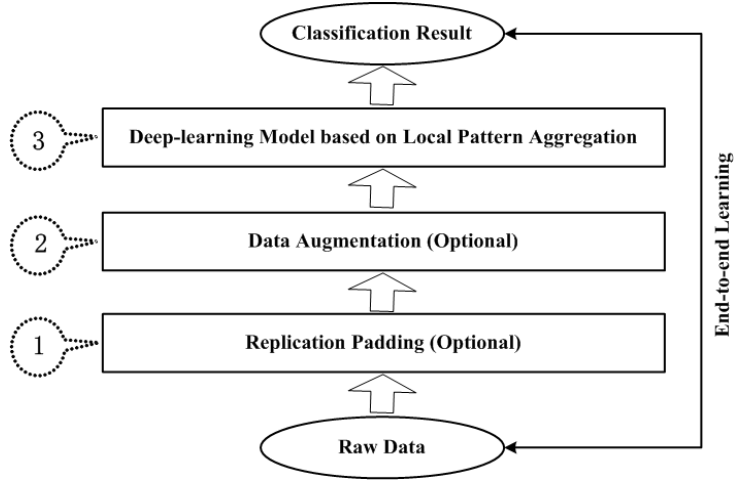

Fig.1 The overall framework of LPANet

1) The raw data such as time series, text sequences, speech, image and video is preprocessed by commonly used methods, for example, filtering, resampling, resizing and normalization, which is then fed into the module (1) to selectively perform the replication padding based on a given application scenario. Using the multivariate time series as an example, the following presents the padding process: let $d_{ij}$ (1<=i<=channelC, 1<=j<=frameC) denote the raw data, padC denote the padding length (padC<=frameC), then the resulting data $d_{ij}'$ can be expressed by

$$d_{ij}' = \begin{cases} d_{ij}, & 1 \leq i \leq channelC,\ 1 \leq j \leq frameC \\ d_{i,j-frameC}, & 1 \leq i \leq channelC,\ frameC < j \leq frameC + padC \end{cases} \quad (1)$$

2) The module (2) takes the raw/replication-padded data as input and selectively performs data augmentation. Similarly, the following describes the processing procedure using multivariate time series as an example. For a given sample, let $d_{ij}$ (1<=i<=channelC, 1<=j<=frameC) denote the raw data, [1, *offset*] denote the range of values allowed for starting point location. Then we pick a time sub-series $d_{ij}'$ expressed by

$$d_{ij}' = d_{i,b-1+j},\ 1 \leq i \leq channelC,\ 1 \leq j \leq frameC - b + 1 \quad (2)$$

where $b \in [1,$ *offset*$]$. After the cropping, $d_{ij}'$ is processed by other data-augmentation strategies, such as stretching or compressing in the time axis by a random factor, contaminating randomly picked segments and adding random noisy/incorrect labels [15]. It is worth noticing that the raw time sub-series is directly fed into the module (3) without any other processing at the testing stage.

3) When accepting the time sub-series, the module (3) produces the probability of belonging to each category and outputs classification results. In the training stage, we assign a random integer in the range of [1, *offset*] to $b$ and train the proposed LPANet in an end-to-end manner using stochastic gradient descent. In the testing stage, we assign one or multiple predefined values to $b$ and average the probability values if necessary. Furthermore, we can generate multiple LPANets using different training strategies and combines their prediction results to achieve better performance.

As for input data such as 2-dimensional images and 3-dimensional videos, we can extend replication padding and cropping to them and adopt other data-augmentation strategies include adding Gaussian noise, rotation, flipping, color altering, etc. [16].

## 2.2 Deep-learning Models based on Local Pattern Aggregation

The core idea behind our proposed LPANet is that the categories of probability for each local pattern (namely local probability value) is implicitly obtained, which are then aggregated to form a global probability value using an application-related formula. Inspired by feature maps in convolutional units, we call the resulting output of such an operation as "classifier map". It is natural to introduce multiple classifier maps, which are fused to get the final probability value. We train the overall model in an end-to-end manner, avoiding the accumulation of errors resulting from intermediate processes.

Our proposed models falls into two types according to the location where crops (corresponding to local patterns) are cut from, "raw data mode" and "transformed features mode". In the former mode, we perform cropping based on the raw data and can directly use any existing deep neural network (DNN) architectures without modification, but the computational complexity for such a scheme is relatively high. In contrast, we pick crops from the transformed features in the latter mode, which means that a given network architectures should be altered. The merit of this scheme is that it can effectively reduce the computational burden. Fig.2 and Fig.3 illustrate the modular frameworks of the proposed model in the raw data and transformed feature modes respectively. The block diagram of DNN in Fig.2 can be a CNN or its variants such as LCNN [17], VGGNet [18] and ResNet [19], but the recurrent structures should not be a component of the DNN. As for local and global aggregation functions, it is the crucial part of the proposed model and we will elaborate on them in the section 2.3.

*A. Raw Data Mode*

There are many ways to pick all crucial crops from given data. A simple yet effective method is as follows: let $d_{ij}$ ($1 \leq i \leq channelC$, $1 \leq j \leq w_2$) denote the input data, then using a sliding window of size $w_1$ with stride $s$, the overlapping crop $t_{ij}^k$ ($1 \leq i \leq channelC$, $1 \leq j \leq w_1$) can be expressed by

$$t_{ij}^k = \begin{cases} d_{i,(k-1)s+j}, & 1 \leq k \leq \lfloor (w_2 - w_1)/s + 1 \rfloor \\ d_{i,w_2-w_1+j}, & k = \lceil (w_2 - w_1)/s + 1 \rceil \end{cases} \quad (3)$$

The last crop can be dropped if ($w_2 - w_1$) is not divisible by $s$. In addition, we can use the followings strategies to perform cropping: (1) Non-overlapping crops are picked from input data. (2) The sizes vary from one crop to another, but the difference in sizes between crops should not be remarkable. (3) A fixed number of crops are picked regardless of input size. (4) The cropping strategy is determined depends on the application scenario, for example, the crop can be heartbeat with different lengths. Note that we should equip deep-learning models with adaptive pooling strategies such as GMP, GAP and SPP if the inputs are variable-size crops.

*B. Transformed Features Mode*

The transformed features mode and the raw data mode are essentially the same; the only difference between them is whether cropping operations are embedded into deep-learning models. Using a 3-layer CNN as an example, Fig.3 presents the internals of the LPANet in the transformed features mode where ConvUnit, FC and LR/SR denote a convolutional unit, a fully-connected layer and a logistic / softmax regression classifier respectively. Evidently, we can use other regression techniques instead of LR/SR, such as fully sigmoid-activated regression used for multi-label classification and ridge regression [20]. The number of hidden layers in a multilayer perceptron is adjusted as needed and the size of the sliding window can be inferred from the kernel size of convolutional units. As shown in Fig.3, "1×2@K3 segment" denotes a 1×2 crop with K3 feature maps, which corresponds to a 600×K input cropped from the raw data.

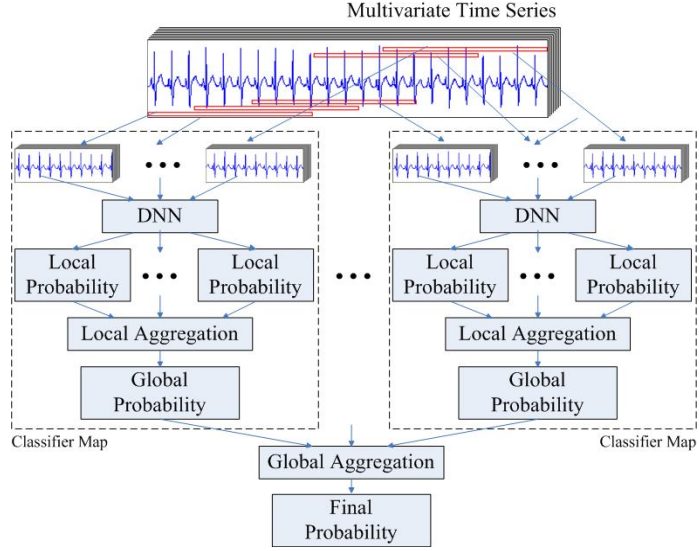

Fig.2 The LPANet in the raw data mode

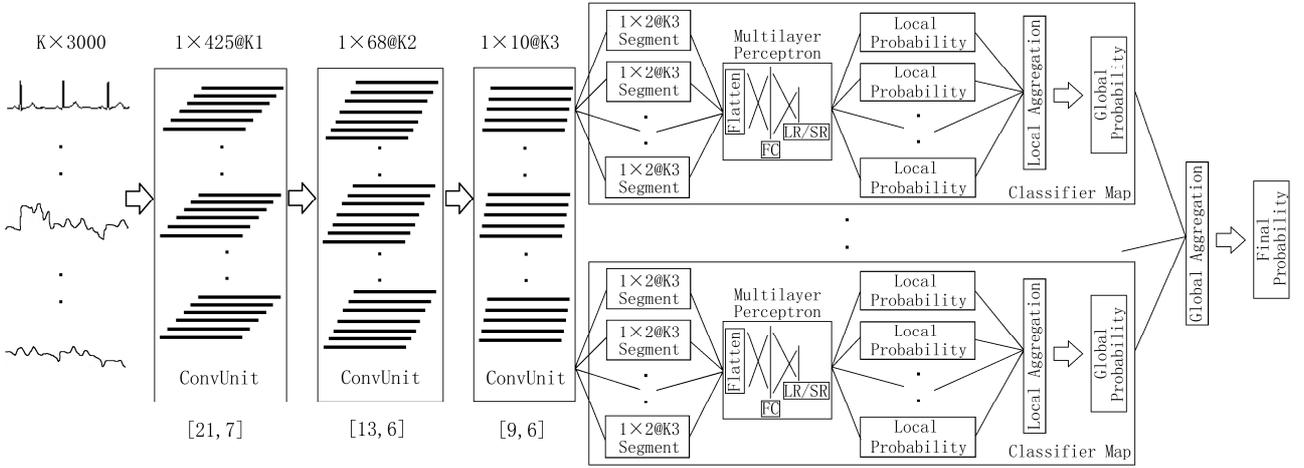

Fig.3 The LPANet in the transformed features mode

## 2.3 Aggregation Function

The core of our proposed model is to aggregate multiple local probability values into a global probability value in each classifier map, which can adopt different cropping strategies mentioned in the section 2.2 or the same one with different parameters, and then to aggregate multiple global probability values into the final probability value. The local and global aggregation functions can be the same or different formulas. Taking one classifier map as an example, the following provides several commonly used aggregation functions in the raw data mode, which can also be suitable for the transformed features mode.

Let $y_1, y_2, \cdots, y_n$ or $\vec{y_1}, \vec{y_2}, \cdots, \vec{y_n}$ be the categories of probability (namely local probability value) for $n$ crops $\vec{x_1}, \vec{x_2}, \cdots, \vec{x_n}$ picked from a sample $\vec{x}$ of the training dataset $DS$, $y$ or $\vec{y}$ be the global probability value (also be the final probability value due to one classifier map), $y_{true}$ or $\vec{y_{true}}$ be the label associated with $\vec{x}$. Denote $G(.)$ as the aggregation function, $F(\vec{x_i}; \Theta)$ as the function mapping from input data to output representation before the regression classifier.

## A. Logistic Regression / Softmax Regression

Logistic regression can only be used for binary classification, and $y_1, y_2, \ldots, y_n$ are all scalar values between 0 and 1 while $y_{true}$ is either 0 or 1, then the probability of belonging to class 1 can be expressed by

$$y_i = \frac{1}{1+\exp\left(-\theta^T F\left(\vec{x}_i;\Theta\right)\right)}, \quad 1 \leq i \leq n \tag{4}$$

Softmax regression is an extension of Logistic Regression used for multi-classification tasks, and $\vec{y}_1, \vec{y}_2, \cdots, \vec{y}_n$ are all vectors while $y_{true}$ is a integer value ranging from 0 to $c$-1 (suppose there are $c$ categories), then we have

$$\begin{cases} \begin{bmatrix} [\vec{y}_1] \\ \vec{y}_2 \\ \vdots \\ \vec{y}_n \end{bmatrix} = \begin{bmatrix} y_1^0 & y_1^1 & \cdots & y_1^{c-1} \\ y_2^0 & y_2^1 & \cdots & y_2^{c-1} \\ \vdots & \vdots & \vdots & \vdots \\ y_n^0 & y_n^1 & \cdots & y_n^{c-1} \end{bmatrix} \\ y_i^j = P(y_{pred} = j \mid \vec{x}_i; \theta) = \dfrac{\exp\left(\theta_j^T F\left(\vec{x}_i;\Theta\right)\right)}{\sum\limits_{k=1}^{c}\exp\left(\theta_k^T F\left(\vec{x}_i;\Theta\right)\right)} \\ \vec{y} = G\left(\vec{y}_1, \vec{y}_2, \cdots, \vec{y}_n\right) = \left[y^0, y^1, \cdots, y^{c-1}\right] \end{cases} \tag{5}$$

where $\theta$ and $\Theta$ are the model parameters needing to be tuned. For Logistic Regression, the output format is the same as that of Softmax Regression used for binary classification if we let $[1-y_i, y_i]$ denote the categories of probability for the $i$-th crop. Hence, we will only elaborate on aggregation functions for Softmax Regression.

Suppose there is an application scenario where the $jm$-th category takes precedence over other categories, then a given sample belongs to that category as long as one crop of it is in the $jm$-th category. For this, we first sort $\vec{y}_1, \vec{y}_2, \cdots, \vec{y}_n$ by $y_i^{jm}$ and obtain $\vec{y}_1^{<jm>}, \vec{y}_2^{<jm>}, \ldots, \vec{y}_n^{<jm>}$ satisfying $y_1^{<jm>} \geq y_2^{<jm>} \geq \ldots \geq y_n^{<jm>}$, and then calculate the final probability value by

$$\vec{y} = \frac{1}{m_2 - m_1 + 1}\sum_{i=m_1}^{m_2} w_i \vec{y}_i^{<jm>}, \quad 1 \leq m_1 \leq m_2 \leq n \tag{6}$$

where $m_1$ and $m_2$ are predefined values that can be adjusted according to the application scenario, and $w_i$ can be either constant or model parameter. Equation (6) corresponds to the simple average method if $w_i = 1$ ($m_1 <= i <= m_2$), and becomes the maximum method if we add an extra condition of $m_1 = m_2 = 1$. When $w_i$ satisfying $\sum\limits_{i=m_1}^{m_2} w_i = m_2 - m_1 + 1$ and $0 \leq w_i \leq m_2 - m_1 + 1$ is tuned by the back propagation algorithm together with other model parameters, equation (6) corresponds to the weighted average method. Besides, we can use the adaptive average method to aggregate local probability values expressed by

$$\vec{y} = \begin{cases} \dfrac{1}{n}\sum\limits_{i=1}^{n}\vec{y}_i, & \text{if } 1\{p_1 \leq y_i^{jm} \leq p_2\} = 0, \ i = 1,2,\cdots,n \\ \dfrac{1}{\sum\limits_{i=1}^{n}1\{p_1 \leq y_i^{jm} \leq p_2\}}\sum\limits_{i=1}^{n}1\{p_1 \leq y_i^{jm} \leq p_2\}\vec{y}_i, & \text{otherwise} \end{cases} \tag{7}$$

where 1{.} is the indicator function so that 1{a true statement} = 1 and 1{a false statement} = 0, $p_1$ and $p_2$ are predefined values that can be adjusted as needed. In addition, we can introduce "majority voting" into the aggregation formula, given by

$$\begin{cases} L^{jm,0} = \{\vec{y_i} \mid y_i^{jm} \leq 0.5,\ 1 \leq i \leq n\} \\ L^{jm,1} = \{\vec{y_i} \mid y_i^{jm} > 0.5,\ 1 \leq i \leq n\} \\ \overrightarrow{t^{jm,k}} = \frac{1}{|L^{jm,k}|} \sum L^{jm,k},\ k = 0,1 \\ km = \arg\max\left(\{|L^{jm,k}| \mid k = 0,1\}\right) \\ \vec{y} = \overrightarrow{t^{jm,km}} \end{cases} \quad (8)$$

If there is more than one prioritized category denoted as the set $IL$ ($2 \leq |IL| \leq c$) in an application scenario, it could result in a race condition and the category corresponding to the maximum probability value is output. For this, we first sort $\vec{y_1}, \vec{y_2}, \cdots, \vec{y_n}$ by $y_i^j (j \in IL)$ respectively, and obtain $\vec{y_1}^{<j>}, \vec{y_2}^{<j>}, \ldots, \vec{y_n}^{<j>}$ that satisfies $y_1^{<j>} \geq y_2^{<j>} \geq \ldots \geq y_n^{<j>}$, and then the aggregation formula corresponding to equation (6) is given by

$$\begin{cases} jm = \arg\max\left(\left\{\frac{1}{m_2 - m_1 + 1} \sum_{i=m_1}^{m_2} w_i y_i^{<j>}, 1 \leq m_1 \leq m_2 \leq n \mid j \in IL\right\}\right) \\ \vec{y} = \frac{1}{m_2 - m_1 + 1} \sum_{i=m_1}^{m_2} w_i \overrightarrow{y_i^{<jm>}},\ 1 \leq m_1 \leq m_2 \leq n \end{cases} \quad (9)$$

The formula corresponding to equation (7) is expressed by

$$\begin{cases} jm = \arg\max\left(\left\{\begin{cases} \frac{1}{n} \sum_{i=1}^{n} y_i^j,\ \text{if } 1\{p_1 \leq y_i^j \leq p_2\} = 0,\ i = 1,2,\cdots,n \\ \frac{1}{\sum_{i=1}^{n} 1\{p_1 \leq y_i^j \leq p_2\}} \sum_{i=1}^{n} 1\{p_1 \leq y_i^j \leq p_2\} y_i^j,\ \text{otherwise} \end{cases} \middle| j \in IL\right\}\right) \\ \vec{y} = \begin{cases} \frac{1}{n} \sum_{i=1}^{n} \vec{y_i},\ \text{if } 1\{p_1 \leq y_i^{jm} \leq p_2\} = 0,\ i = 1,2,\cdots,n \\ \frac{1}{\sum_{i=1}^{n} 1\{p_1 \leq y_i^{jm} \leq p_2\}} \sum_{i=1}^{n} 1\{p_1 \leq y_i^{jm} \leq p_2\} \vec{y_i},\ \text{otherwise} \end{cases} \end{cases} \quad (10)$$

The formula corresponding to equation (8) is given by

$$\begin{cases} L^j = \{\vec{y_i} \mid y_i^j > 0.5,\ 1 \le i \le n\}, j \in IL \\ \vec{t^j} = \dfrac{1}{|L^j|} \sum L^j \\ jm = \arg\max\left(\{|L^j| \mid j \in IL\}\right) \\ \vec{y} = \begin{cases} \dfrac{1}{n}\sum\limits_{i=1}^{n}\vec{y_i}, & \text{if } L^j = \phi, j \in IL \\ \vec{t^{jm}}, & \text{otherwise} \end{cases} \end{cases} \quad (11)$$

Beside the above mentioned equations, we can use any linear and non-linear aggregation formulas as long as they can make the loss function, i.e., equation (12) differentiable with respect to model parameters

$$\text{cost} = -\sum_{(\vec{x}, y_{true}) \in DS} \sum_{j=0}^{c-1} 1\{y_{true} = j\} \log(y^j) \quad (12)$$

For instance, we can introduce attentional mechanisms into aggregation functions, and the attentional model with one hidden layer and a single output is expressed by

$$\begin{cases} a_i = \dfrac{\exp\left(W_2^T \sigma\left(W_1^T F(\vec{x_i};\Theta) + b\right)\right)}{\sum\limits_{i=1}^{n} \exp\left(W_2^T \sigma\left(W_1^T F(\vec{x_i};\Theta) + b\right)\right)}, 1 \le i \le n \\ \vec{y} = \sum\limits_{i=1}^{n} a_i \vec{y_i} \end{cases} \quad (13)$$

where $W_2, W_1$ and $b$ are all model parameters needing to be tuned, and $a_i$ is a scalar value. We can also modify equation (13) to output attentional weight $\vec{a_i}$ with the same length as vector $\vec{y_i}$. Moreover, "skip summation" and "nonlinear summation" can be introduced into aggregation functions, and the following provides some typical examples based on the simple average method, including Noisy-or [21], Int-seg-rec [22], Generalized Mean [23], Log-sum-exp [23], Noisy-and [24], Linear Softmax [25], Exp. Softmax [25] and Even-add.

$$\vec{y} = 1 - \prod_{i=m_1}^{m_2} (1 - \vec{y_i}),\ 1 \le m_1 \le m_2 \le n \quad (14)$$

$$\vec{y} = \dfrac{\sum\limits_{i=m_1}^{m_2} \dfrac{\vec{y_i}}{1 - \vec{y_i}}}{1 + \sum\limits_{i=m_1}^{m_2} \dfrac{\vec{y_i}}{1 - \vec{y_i}}},\ 1 \le m_1 \le m_2 \le n \quad (15)$$

$$\vec{y} = \left(\dfrac{1}{m_2 - m_1 + 1} \sum_{i=m_1}^{m_2} (\vec{y_i})^r\right)^{\frac{1}{r}},\ 1 \le m_1 \le m_2 \le n \quad (16)$$

$$\vec{y} = \dfrac{1}{r} \log\left(\dfrac{1}{m_2 - m_1 + 1} \sum_{i=m_1}^{m_2} e^{r\vec{y_i}}\right),\ 1 \le m_1 \le m_2 \le n \quad (17)$$

$$\vec{y} = \frac{\sigma\left(a\left(\frac{1}{m_2 - m_1 + 1}\sum_{i=m_1}^{m_2}\vec{y}_i - b\right)\right) - \sigma(-ab)}{\sigma(a(1-b)) - \sigma(-ab)}, \ 1 \leq m_1 \leq m_2 \leq n \tag{18}$$

$$\vec{y} = \frac{\sum_{i=m_1}^{m_2}\vec{y}_i^{\,2}}{\sum_{i=m_1}^{m_2}\vec{y}_i}, \ 1 \leq m_1 \leq m_2 \leq n \tag{19}$$

$$\vec{y} = \frac{\sum_{i=m_1}^{m_2}\vec{y}_i\exp(\vec{y}_i)}{\sum_{i=m_1}^{m_2}\exp(\vec{y}_i)}, \ 1 \leq m_1 \leq m_2 \leq n \tag{20}$$

$$\vec{y} = \frac{1}{\lfloor(m_2 - m_1)/2\rfloor + 1}\sum_{i=0}^{\lfloor(m_2 - m_1)/2\rfloor}\vec{y}_{m_1 + 2i}, \ 1 \leq m_1 \leq m_2 \leq n \tag{21}$$

where $r$ in equation (16) and (17) can be either constant or model parameter, and $\sigma$ in equation (18) is a sigmoid function, $a$ is a predefined value whereas $b$ is a model parameter in the range of $[0,1]$. Note that equation (14) ~ (21) require local probability values to be sorted first, just as we do earlier, and can be directly used only if $\vec{y}_i$ is scalar. The calculation is slightly different when $\vec{y}_i$ is a vector, and the following describes the details taking equation (14) as example. Let $IL$ ($1 <= |IL| <= c$) denote the set of prioritized categories, and $NIL$ ($|IL| + |NIL| = c$) denote the set of non-prioritized categories, then we have

$$\begin{cases} \widetilde{y^j} = 1 - \prod_{i=m_1}^{m_2}(1 - y_i^j), \ j \in IL \\ \widetilde{y^j} = \frac{1}{m_2 - m_1 + 1}\sum_{i=m_1}^{m_2}y_i^j \ \text{or} \ y_{m_1}^j \ \text{or} \ y_{m_2}^j, \ j \in NIL \\ y^j = \frac{\widetilde{y^j}}{\sum_{k=0}^{c-1}\widetilde{y^k}}, \ j = 0, 1, \cdots, c-1 \\ \vec{y} = [y^0, y^1, \cdots, y^{c-1}] \end{cases} \tag{22}$$

As the no free lunch theorem [26] suggests, there is not an algorithm that always work effectively on all problems, thus we should choose or design the best aggregation formula for a given pattern recognition task.

### B. Fully Sigmoid-activated Regression

Fully sigmoid-activated regression can be used for multi-label classification, and $\vec{y}_1, \vec{y}_2, \cdots, \vec{y}_n, \vec{y}_{true}$ are all vectors where $\vec{y}_{true} = [y_{true}^0, y_{true}^1, \cdots, y_{true}^{c-1}]$ and $y_{true}^j$ is either 0 or 1 (suppose there are $c$ categories), then we have

$$\begin{cases} \begin{bmatrix} \overrightarrow{y_1} \\ \overrightarrow{y_2} \\ \vdots \\ \overrightarrow{y_n} \end{bmatrix} = \begin{bmatrix} y_1^0 & y_1^1 & \cdots & y_1^{c-1} \\ y_2^0 & y_2^1 & \cdots & y_2^{c-1} \\ \vdots & \vdots & \vdots & \vdots \\ y_n^0 & y_n^1 & \cdots & y_n^{c-1} \end{bmatrix} \\ y_i^j = P(y_{pred} = j \mid \overrightarrow{x_i}; \theta) = \dfrac{1}{1 + \exp\left(\theta_{j+1}^T F\left(\overrightarrow{x_i}; \Theta\right)\right)} \\ \overrightarrow{y} = G\left(\overrightarrow{y_1}, \overrightarrow{y_2}, \cdots, \overrightarrow{y_n}\right) = \left[y^0, y^1, \cdots, y^{c-1}\right] \end{cases} \qquad (23)$$

where $\theta$ and $\Theta$ are the model parameters needing to be tuned. The aggregation functions for this type of regression technology are almost the same as that of Softmax Regression; the only difference is that there is neither a prioritized category, nor any other rival category. Since all categories are equal, we aggregate local probability values for each category independently. Specifically, we first sort $\overrightarrow{y_1}, \overrightarrow{y_2}, \cdots, \overrightarrow{y_n}$ by $y_i^j (0 \leq j \leq c-1)$ respectively and obtain $\overrightarrow{y_1^{<j>}}, \overrightarrow{y_2^{<j>}}, \ldots, \overrightarrow{y_n^{<j>}}$ satisfying $y_1^{<j>} \geq y_2^{<j>} \geq \ldots \geq y_n^{<j>}$, and then the aggregation formula corresponding to equation (6) is expressed by

$$y^j = \dfrac{1}{m_2 - m_1 + 1} \sum_{i=m_1}^{m_2} w_i y_i^{<j>}, \quad 1 \leq m_1 \leq m_2 \leq n \qquad (24)$$

The formula corresponding to equation (7) is given by

$$y^j = \begin{cases} \dfrac{1}{n} \sum_{i=1}^{n} y_i^j, & \text{if } 1\{p_1 \leq y_i^j \leq p_2\} = 0, \ i = 1, 2, \cdots, n \\ \dfrac{1}{\sum_{i=1}^{n} 1\{p_1 \leq y_i^j \leq p_2\}} \sum_{i=1}^{n} 1\{p_1 \leq y_i^j \leq p_2\} y_i^j, & \text{otherwise} \end{cases} \qquad (25)$$

The formula corresponding to equation (8) is expressed by

$$\begin{cases} L^{j,0} = \left\{ y_i^j \mid y_i^j \leq 0.5, \ 1 \leq i \leq n \right\} \\ L^{j,1} = \left\{ y_i^j \mid y_i^j > 0.5, \ 1 \leq i \leq n \right\} \\ t^{j,k} = \dfrac{1}{|L^{j,k}|} \sum L^{j,k}, \ k = 0, 1 \\ km = \arg\max\left(\left\{ |L^{j,k}| \mid k = 0, 1 \right\}\right) \\ y^j = t^{j,km} \end{cases} \qquad (26)$$

Similarly, we can use any linear and non-linear aggregation formulas as long as they can make the loss function, i.e., equation (27) differentiable with respect to model parameters

$$\text{cost} = -\sum_{(\overrightarrow{x}, y_{true}) \in DS} \sum_{j=0}^{c-1} \left[ y_{true}^j \log(y^j) + (1 - y_{true}^j) \log(1 - y^j) \right] \qquad (27)$$

In addition, we can add extra regularization terms based on the relationship between categories in equation (27) to achieve better performance.

*C. Recurrent Regression*

Equation (4), (5) and (23) assume that local probability values are weakly dependent on each other, and cannot be used in the situation where strong dependency exist between different crops, such as natural language processing [27]. To overcome this issue, we introduce recurrent regression into our proposed LPANet. Suppose there are $c$ output nodes (i.e., there are $c$ categories), then we have

$$\begin{cases} \vec{y}_i = \sigma\left(W\vec{y}_{i-1} + \theta^T F(\vec{x}_i; \Theta)\right), 1 \leq i \leq n \\ \vec{y} = G(\vec{y}_1, \vec{y}_2, \cdots, \vec{y}_n) = [y^0, y^1, \cdots, y^{c-1}] \end{cases} \quad (28)$$

where $W$, $\theta$ and $\Theta$ are the model parameters needing to be tuned. The activation function $\sigma$ can be the logistic regression function with both $\vec{y}_i$ and $W$ being scalar if $c$ is equal to 1; otherwise, it can be the softmax regression (or fully sigmoid-activated regression) function with $\vec{y}_i$ being a vector, and $W$ is either scalar, vector or matrix. However, too many crops could result in unstable gradients when using the back propagation algorithm, thus we can introduce gating mechanisms [11-12] or transformer [28] into equation (28). As for global probability value $\vec{y}$, we can use the above mentioned aggregation formulas to calculate it or directly assign $\vec{y}_n$ to it. Note that it is not recommended to use this type of regression technology due to its difficult parameter optimization problem and high computational complexity, unless neighboring crops are strong dependent.

# 3. Case Study

In this section, we will apply the proposed LPANet to a typical pattern recognition task, namely premature ventricular contraction (PVC) detection using single-lead short ECG. We will first present the overall architecture and implementation details, and then provide the experimental results as well as comparison with some well-known methods.

## 3.1 Detailed Implementation

As a common form of cardiac arrhythmia, PVC can lead to life-threatening cardiac conditions. Accurate and timely detection of PVC has a great clinical significance. Computer-aided PVC detection would be beneficial to reduce the clinical workload of physicians, thus many methods have been proposed for this subject, including feature extraction-based and deep learning-based approaches. Among these work, CNN and LSTM-based methods [29] achieve excellent results but suffer from high computational complexity, and rules inference-based methods [30-31] may be a good choice but the involved parameters are sensitive to different testing sets. To overcome the issue, our proposed LPANet is utilized for PVC detection.

We first resample a raw ECG recording to sampling frequency 150 Hz, and then filter it with a bandpass filter with a passband from 0.5Hz to 50Hz. Finally, we pad the processed data to a fixed length (20 seconds) with a random signal whose amplitudes vary from 0 to 0.1, and thus the resulting ECG recording has a size of $1 \times 3000$. For simplicity's sake, we only choose lead II signal and do not adopt any data-augmentation strategies in the evaluation experiment, meaning that $b$ in equation (2) is always equal to 1. The reason why we pad ECG recordings with noisy signals is to evaluate the robustness of PVC detectors.

We employ the LPANet with one classifier map in the raw data mode (shown in Fig.2) and used a 3-layer one-dimensional CNN (1DCNN, similar to Fig.3 with the cropping and aggregation operations removed) as its DNN component. We adopt equation (3) with $s = 257$ to pick $1\times1200$ overlapping segments from an incoming ECG recording. The total number of segments is 8 as the last one is dropped. The CNN accepts $1\times1200$ input data, utilizing convolutional units (ConvUnit) to extract features and a multilayer perceptron composed of a fully-connected layer with 50 nodes and a logistic regression layer to calculate the categories of probability. The kernel sizes of three ConvUnits are [21, 7], [13, 6] and [9, 6] respectively (21 denotes the convolutional kernel size, and 7 denotes the kernel size for the pooling layer). The numbers of feature maps in the three ConvUnits are 6, 7 and 5 separately. The problem of PVC detection is a typical binary classification task and there are only two categories, PVC and non-PVC. We set PVC as the prioritized category and use equation (6) with $m_1 = m_2 = 1$ and $w_1 = 1$ for aggregation purpose. The overall LPANet is trained in an end-to-end manner using stochastic gradient descent based on Adam's update rule [32]. The goal is to minimize the loss function, i.e., equation (12) by tuning model parameters. Our network model runs on the popular deep learning framework Keras-2.2.4 [33] using Theano-1.04 [34] backend. The number of epochs and the mini-batch size are set to 100 and 256 respectively. To solve the class imbalance problem, we replicate the PVC recordings about $23.8(\approx10423/438)$ times in the training set. As for the parameters of Adam optimizer, we select the default values set in Keras.

We also employ the classical 1DCNN and 1DCNN+LSTM models for PVC detection. To make a fair comparison, the training strategies and the values of hyper-parameters remain the same. The key differences are that the input size of both 1DCNN and 1DCNN+LSTM is $1\times3000$ and the fully-connected layer with 50 nodes is replaced with a LSTM layer in the latter one.

## 3.2 Results

The Aliyun-Tianchi ECG Challenge Database 2019 (AliyunDB2019) [35] is composed of 32,142 eight-lead (i.e., lead I, II, and V1~V6) ECG recordings, each digitized at 500Hz with 10 seconds in duration. Only 24,106 of them have diagnostic conclusions, including 55 types of diseases such as ventricular premature beat, atrial fibrillation and right bundle branch block. The training dataset consists of 438 PVC recordings and 10,423 Non-PVC recordings, while the testing dataset is composed of 485 PVC recordings and 11,562 Non-PVC recordings. We also create a validation dataset for model selection during the training phase, which contains 48 PVC recordings and 1,150 Non-PVC recordings. Here PVC only denotes premature ventricular contraction, and Non-PVC includes all other cardiac rhythms, not just sinus rhythm. We utilize several metrics to investigate the PVC detection performance, including sensitivity (*Se*), specificity (*Sp*), accuracy (*Acc*) and F-score ($F_{PVC}$, $F_{Non-PVC}$ and $F_{AVG}$). Table 1 illustrates the experimental results of our proposed LPANet as well as comparison with the 1DCNN and 1DCNN+LSTM.

Table 1 Detection performance comparisons of PVC versus other rhythms

| Model | *Se* | *Sp* | *Acc* | $F_{PVC}$ | $F_{Non-PVC}$ | $F_{AVG}$ |
|---|---|---|---|---|---|---|
| LPANet | 65.15% | **98.26%** | **96.93%** | **63.07%** | **98.40%** | **80.74%** |
| 1DCNN | 62.47% | 96.84% | 95.46% | 52.56% | 97.62% | 75.09% |
| 1DCNN+LSTM | 66.80% | 97.70% | 96.46% | 60.28% | 98.15% | 79.21% |

As we can see from Table 1, our proposed LPANet outperforms the other two network models in almost all the metrics. Due to the noisy signals introduced in ECG recordings, the 1DCNN fails to recognize the discriminated features and tends to misclassify many input ECG recordings. It leads to low classification metrics in detecting PVC recordings. Thanks to the powerful ability for time series modeling, the 1DCNN+LSTM reaches 4.33% increment in sensitivity, 0.86% increment in specificity, 1.00% increment in accuracy and 4.12% increment in F-score, compared to the 1DCNN. However, when a part of input ECG recording is a noisy signal, it is vital to ignore details extracted from the corresponding region. This also means that different ECG segments would be better to be weakly dependent on each other. Our proposed LPANet therefore outperforms the 1DCNN+LSTM in 5 out of the 6 metrics, improving the PVC detection performance. To summarize, the experimental results demonstrate the potential and effectiveness of the proposed LPANet.

## 4. Discussion

It is quite common to accept variable-size input data for deep CNNs in practical use, but they require a fixed input size. For this, simple data preprocessing methods such as cropping, padding and warping, and adaptive pooling strategies such as GMP, GAP and SPP are utilized to remove the restriction. However, the performance tends to be degraded when input sizes vary greatly. A preferable method is to introduce recurrent structures such as LSTM and GRU, but it suffers from high computational complexity and increases the difficulty of turning model parameters.

Multiple instances learning (MIL) originally proposed for drug activity prediction [36] is a kind of weakly supervised learning. The input data is a set of labeled bags, each containing multiple instances. Formally, there is a dataset $D = \{(X_i, Y_i)\}$, $i = 1, 2, \ldots, n$, where $X_i = \{x_{i,j}\}$, $j = 1, 2, \ldots, m_i$, $Y_i \in \{0, 1\}$ is the label of the $i$-th bag $X_i$, $n$ and $m_i$ denote the number of bags and the number of instances in bag $X_i$ respectively. Let $y_{ij} \in \{0, 1\}$ denote the label of instance $x_{i,j}$, then $Y_i = 0$ if and only if all $y_{ij} = 0$; otherwise, $Y_i = 1$. Note that only bag label $Y_i$ is given and instance label $y_{ij}$ remains unknown during training. Many algorithms have been proposed to solve the MIL problem. Due to the great successes of deep learning, the solution using neural networks [23, 37-40] draw more attention. The multiple instance neural network (MINN) takes bags with a variable number of instances as input and directly output the bag labels. The model parameters of a MINN can be tuned using stochastic gradient descent in an end-to-end manner.

General speaking, the MIL problem mainly refers to a classification task and a bag of instances exhibit neither dependency nor ordering among each other [41-43]. Although some research works [24, 44-46] have utilized MINNs to process variable-size input data, they subject to the above constraints from beginning to end. Our proposed LPANet is somewhat similar to MINN, but it greatly expands the scope and features available in traditional MIL. By introducing prioritized categories and rival categories, MIL can now be used for multi-classification tasks in a different way. The classifier map bridges MIL and ensemble learning, and recurrent regression bridges MIL and recurrent neural network. By combing adaptive pool and recurrent regression, we can implement the concept of variable-size time step. To summarize, LPANet fuses local pattern aggregation, deep learning, multiple instances learning and selective ensemble learning into a single learning body, and solve the issue involved with recurrent structures to some extent. Another important contribution is to use the replication-padding strategy together with LPANet. For a given sample in some application scenarios, the most discriminated features could be located at the boundary regions. Without the replication-padding operation, randomly picked crops may not contain discriminated features, thus the performance tends to be degraded.

## 5. Conclusion

The LPANet is a flexible solution for handling variable input sizes. This issue is important in time series prediction, text classification, speech recognition, object detection, and so on. The suggested solution greatly extends traditional MIL, and reduces the difficulty of tuning model parameters. Using a case study concerning PVC detection, the potential and effectiveness of the LPANet is validated. Our PVC detector has better performance on the AliyunDB2019 compared with classical deep-learning models. In our future work, we will present more case studies to demonstrate the superiority of the proposed model.